\documentclass[10pt,a4paper,oneside,fleqn]{article}
\usepackage[a4paper,left=1.5cm,right=1.5cm,top=0.65cm,bottom=1.3cm,includeheadfoot]{geometry}
\usepackage{mathrsfs}
\usepackage{amsmath}
\usepackage{amsfonts}
\usepackage{multicol}
\usepackage{graphicx}
\usepackage{CJK}
\usepackage[numbers,square,sort&compress]{natbib}
\usepackage{multirow}
\usepackage{booktabs}
\usepackage{float}
\usepackage{afterpage}
\usepackage{dblfloatfix}
\usepackage{fancyhdr}
\usepackage{graphics}
\usepackage{tabularx}
\usepackage{algorithm}
\usepackage{algpseudocode}
\usepackage{caption}
\usepackage[table]{xcolor}
\newcommand{\R}{\mathbb{R}}
\usepackage[colorlinks=true, linkcolor=blue, citecolor=blue, urlcolor=blue]{hyperref}
\newcommand{\secref}[2]{\textcolor{black}{Sec.}~\hyperref[#1]{\textcolor{blue}{#2}}}

\catcode`@=12  
\begin{document}

\twocolumn[
\begin{@twocolumnfalse}
\noindent {\bf\LARGE{Compressive sensing inspired self-supervised \\ single-pixel imaging}}\vspace*{8mm}

\noindent {\bf\normalsize{Jijun Lu,$^{\rm a,b,\dagger}$ Yifan Chen,$^{\rm a,c,\dagger}$ Libang Chen,$^{\rm a}$ Yiqiang Zhou,$^{\rm a}$ Ye Zheng,$^{\rm a}$ Mingliang Chen,$^{\rm d}$ \\
Zhe Sun,$^{\rm a,e,f,\ast}$ and Xuelong Li$^{\rm a, \ast}$}}

\noindent{\footnotesize{$^{\rm a}$Institute of Artificial Intelligence (TeleAI), China Telecom, China}}\\
\noindent{\footnotesize{$^{\rm b}$School of Integrated Circuits, Shanghai Jiao Tong University, Shanghai, China}}\\
\noindent{\footnotesize{$^{\rm c}$College of Future Information Technology, Fudan University, Shanghai, China}}\\
\noindent{\footnotesize{$^{\rm d}$Aerospace Laser Technology and System Department, Shanghai Institute of Optics and Fine Mechanics, Chinese Academy of Sciences, Shanghai, China}}\\
\noindent{\footnotesize{$^{\rm e}$School of Artificial Intelligence, Optics and Electronics (iOPEN), Northwestern Polytechnical University, Xi'an, China}}\\
\noindent{\footnotesize{$^{\rm f}$Fujian Ocean Innovation Center, Xiamen, China}}\\

\noindent{\small{\textbf{Abstract.} Single-pixel imaging (SPI) is a promising imaging modality with distinctive advantages in strongly perturbed environments. Existing methods lack physical sparsity constraints and overlook the local-global feature integration, leading to severe noise vulnerability, structural distortions and blurred details. To address these limitations, we propose SISTA-Net, a compressive sensing-inspired self-supervised method for SPI. SISTA-Net adopts an optimization-aligned architecture based on the Iterative Shrinkage-Thresholding Algorithm (ISTA), comprising a data fidelity module and a proximal mapping module. First, the fidelity module adopts a hybrid CNN-Visual State Space Model (VSSM) architecture to integrate local and global feature modeling, enhancing reconstruction integrity and fidelity. Second, we leverage deep nonlinear networks as adaptive sparse transforms combined with a learnable soft-thresholding operator to impose explicit physical sparsity in the latent domain, enabling noise suppression and robustness to interference even at extremely low sampling rates. Extensive experiments on multiple simulation scenarios demonstrate that SISTA-Net outperforms state-of-the-art methods by 2.6~dB in PSNR. Real-world far-field underwater tests yield a 3.4~dB average PSNR improvement, validating its robust anti-interference capability.}}\vspace*{2mm}
  
\noindent{\small{\textbf{Keywords: }single-pixel imaging; compressive sensing; deep learning. }}\vspace*{2mm}

\noindent{\small{$^\ast$ Address all correspondence to Zhe Sun, sunzhe@nwpu.edu.cn; Xuelong Li, xuelong\_li@ieee.org}}\\
\noindent{\small{$^\dagger$ These authors contributed equally to this work.}}\vspace*{10mm}
\end{@twocolumnfalse}
]
\sloppy{}
\normalsize

\noindent\textbf{1\quad Introduction}\vspace*{2mm}

\noindent Single-pixel imaging (SPI) has emerged as a promising computational imaging technique. As a non-local imaging modality based on the correlation properties of second-order spatiotemporal intensity fluctuations of light fields\cite{wang2025deep}, it offers distinctive advantages in strongly perturbed environments, such as highly scattering media\cite{o-2026-underwater-imaging,2024multi-polarization-GI} or low-light conditions\cite{edgar2019principles}. Early computational SPI heavily relied on correlation-based physical model approaches\cite{shapiro2008computational}, thus requiring massive optical measurements and leading to prohibitive acquisition times. Accordingly, compressive sensing (CS) techniques\cite{baraniuk2007compressiveCS,katz2009compressiveGI} were introduced into SPI\cite{duarte2008single-pixel-CS}, significantly reducing the required measurements and enabling effective recovery at sub-Nyquist sampling rates\cite{katkovnik2012compressive}. However, these CS-based methods hardly ensure stable reconstruction quality, since conventional linear transformations struggle to capture sparse priors of natural images.\par
Deep learning has made substantial strides in SPI reconstruction\cite{shimobaba2018computational,o-2025largemodel-GI,chen2025compressive}. For example, Lyu et al.\cite{lyu2017deep} introduced a deep learning-based ghost imaging method, leveraging neural networks to map traditionally reconstructed images to their corresponding targets. By leveraging such data-driven architectures\cite{yang2021underwater}, deep learning methods offer key advantages including faster reconstruction speeds\cite{rizvi2020deepghost, kataoka2022noise} and superior imaging quality\cite{wang2019learning}. Despite these advancements, most existing deep neural networks operate as ``black boxes''\cite{lecun2015deep}, lacking interpretability. Furthermore, they are prone to noise artifacts and exhibit degraded reconstruction performance in high-disturbance environments. To address these limitations, deep unrolling networks have emerged to bridge the gap between traditional optimization algorithms and deep learning\cite{gregor2010learning, wang2025proximal}. Zhang et al.\cite{ISTA-Net} unrolled the iterative optimization algorithms into a structured deep network to solve the image inverse problem, effectively integrating CS models into network architectures.\par

\begin{figure*}[htbp]
\centering
\includegraphics[width=0.98\textwidth]{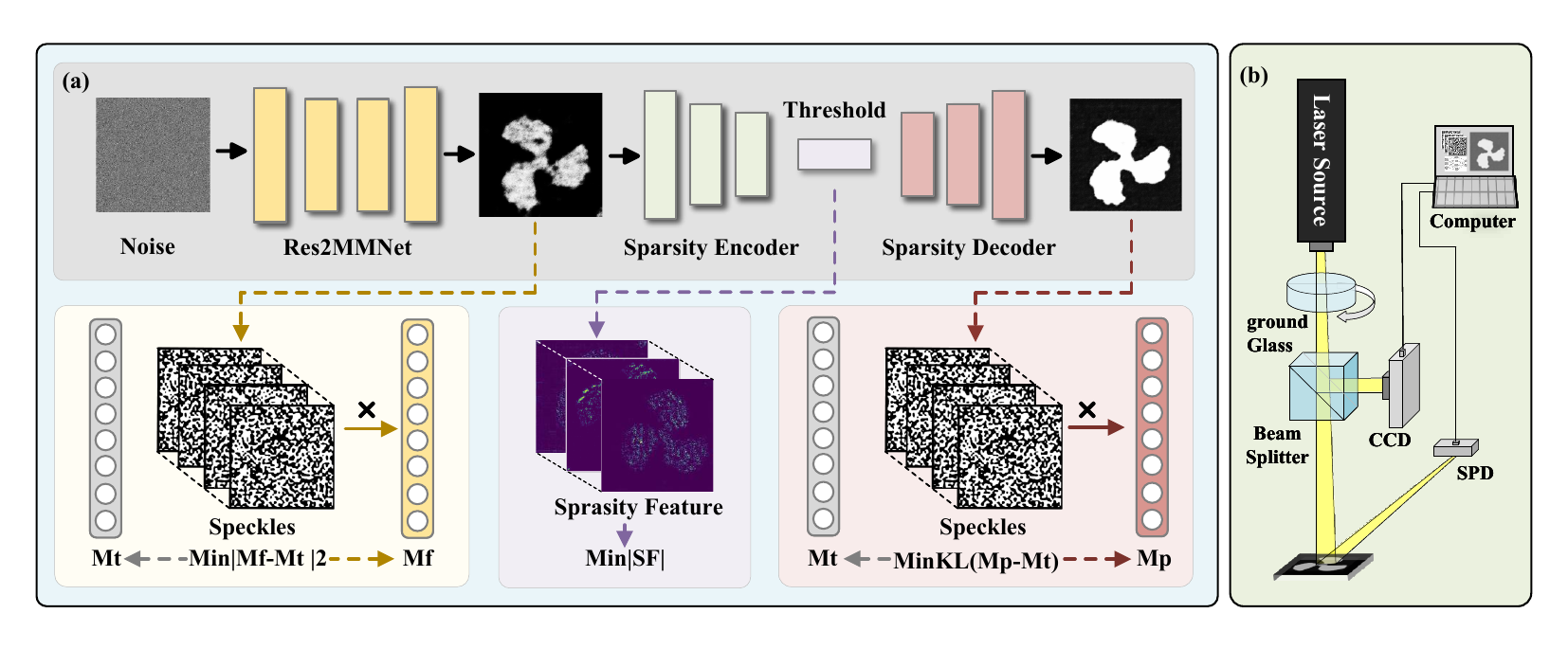}
\caption{Overview of the proposed method: (a) Core reconstruction workflow of the SISTA-Net algorithm. (b) Experimental setup for single-pixel imaging.}
\label{fig:pipeline}
\end{figure*}

\noindent Although these deep unrolling networks successfully leverage sparse priors\cite{TansCS,ADMMGI}, they, like conventional deep learning, suffer from severe data dependency. The requirement for large paired datasets is impractical in real-world scenarios, limiting their generalizability. To overcome the data-driven constraint, self-supervised learning strategies have been increasingly adopted to learn robust feature representations directly from measurement data \cite{o-2024multi-input-GI,zhang2023vgennet}. For instance, Wang et al. \cite{GIDC-GI} utilized 1D measurements as output constraints by integrating the physical model \cite{gibson2020single} into deep learning, thereby effectively eliminating the reliance on paired data. However, although deep unfolding networks and self-supervised learning strategies have respectively addressed the lack of sparsity constraints and the data reliance, there still lacks a unified framework that enforces sparsity for noise suppression within a completely dataset-free paradigm. Furthermore, constrained by the inherent trade-off between convolution-based\cite{wang2021single} local extraction and attention-driven\cite{o-2025attention-GI, qu2024dual} global modeling, current architectures struggle to concurrently capture local texture details and establish global spatial correlations.\par
In this paper, we propose a novel Compressive Sensing-Inspired Self-Supervised Single-Pixel Imaging method (SISTA-Net). Leveraging the Deep Image Prior (DIP) paradigm, our approach requires only 1D intensity measurements as supervisory signals. Guided by the Iterative Shrinkage-Thresholding Algorithm (ISTA), the network is decoupled into a fidelity module and a proximal mapping module. To address the aforementioned representation gap, the fidelity module employs a hybrid CNN-Visual State Space Model (VSSM) architecture to simultaneously extract local fine-grained patterns and model global long-range interactions. The proximal mapping module adopts a sparsity encoder-decoder and a learnable soft-thresholding operator to perform noise filtering in the latent feature domain, enabling key structure preservation even at extremely low sampling rates and enhancing the model's anti-interference capability. These designs enable our method to achieve high-fidelity imaging while improving mathematical interpretability.\par

\begin{figure*}[!h]
\centering
\includegraphics[width=0.98\textwidth]{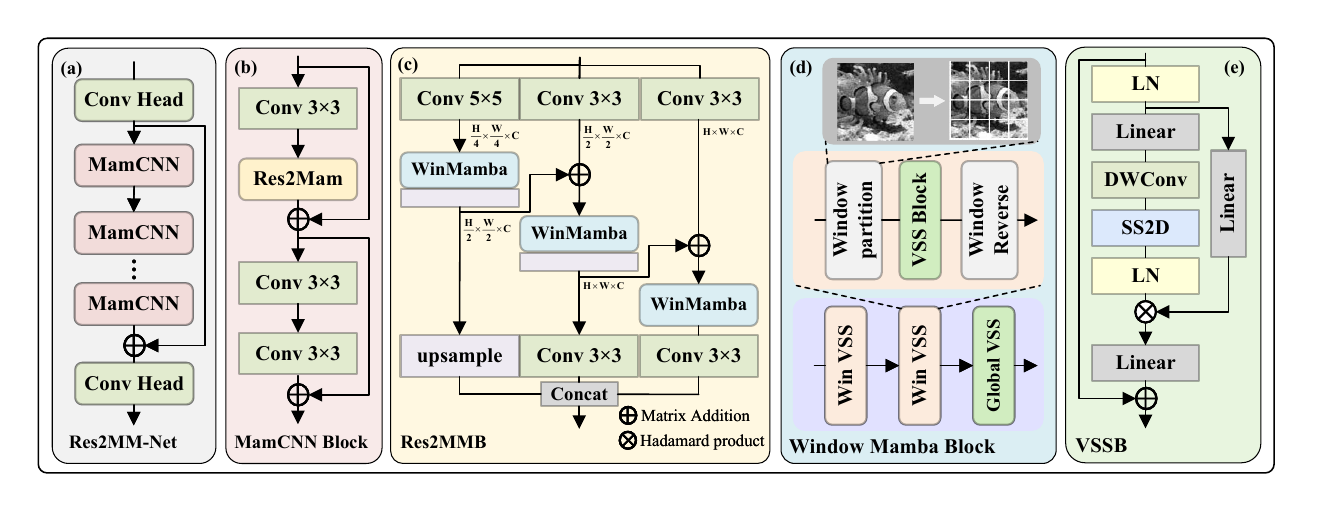}
\caption{Architecture of the Data Fidelity Module. (a) The architecture of Res2MM-Net. (b) The structure of MamCNN Block. (c) The structure of Res2Mam Multi-Scale Block. (d) The structure of Window Mamba Block. (e) The structure of the Vision State Space Block.}
\label{fig:Net1}
\end{figure*}

\vspace*{2mm} \noindent \textbf{2\quad Methodology}
\vspace*{2mm}

\noindent\emph{2.1\quad Proposed Method}\label{sec:proposed_method}\vspace*{2mm}

\noindent In this section, we first provide a brief review of the traditional ISTA for compressed sensing-based SPI, followed by the architecture of the proposed SISTA-Net.

\noindent\textbf{Traditional ISTA}: Optimization-based compressed sensing algorithms\cite{donoho2006compressed} aim to recover the original object image $\mathbf{x}$ from compressed measurements $\mathbf{y} = \mathbf{\Phi}\mathbf{x} + \mathbf{e}$ in SPI systems, where $\mathbf{\Phi}$ denotes the sensing matrix matching SPI's spatial light modulation, $\mathbf{e}$ is additive noise, and this inverse problem is ill-posed\cite{bertero1988ill}. For CS-based SPI, this is formulated as a constrained minimization problem balancing data fidelity and sparsity regularization:  
\begin{equation}
\mathbf{x}^* = \arg\min_{\mathbf{x}} \frac{1}{2} \|\mathbf{\Phi} \mathbf{x} - \mathbf{y}\|_2^2 + \lambda \|\mathbf{\Psi} \mathbf{x}\|_1,
\end{equation}
where $\mathbf{\Psi}$ is the sparsifying operator (e.g., wavelet transform) and $\lambda$ controls the trade-off between measurement consistency (first term) and sparsity enforcement (second term). ISTA\cite{daubechies2004iterative} solves this via two alternating steps: gradient descent for data consistency with $\mathbf{z}^{k+1} = \mathbf{x}^k - \eta \mathbf{\Phi}^\top (\mathbf{\Phi} \mathbf{x}^k - \mathbf{y})$ ($\eta$ is the step size), and proximal mapping with soft-thresholding to enforce sparsity with $\mathbf{x}^{k+1} = \text{sgn}(\mathbf{z}^{k+1}) \max(|\mathbf{z}^{k+1}| - \lambda\eta, 0)$. These two steps are repeated until convergence for SPI reconstruction.\par

Inspired by ISTA, SISTA-Net organizes the reconstruction process into two physics-informed components: a fidelity module and a proximal mapping module, as illustrated in Fig.~\ref{fig:pipeline}(a). The network adopts the Deep Image Prior (DIP) strategy \cite{ulyanov2018deep}, utilizing a fixed random noise tensor $I$ as the initial input\cite{o-2023-turbulent-GI} to reconstruct target images under the supervision of 1D measurements $\mathbf{Y}$ to avoid reliance on paired datasets.\par

In the first stage, the noise tensor $I$ is fed into the fidelity module (described in \secref{sec:fidelity_module}{2.2}) with learnable parameters $\theta_f$:
\begin{equation}
O_F = F(I; \theta_f).
\end{equation}
This forward mapping is designed to play a data-consistency role analogous to the gradient-descent step in ISTA. By learning this mapping, the fidelity module acts as an implicit data consistency solver, ensuring that the intermediate output $O_F$ captures structural information consistent with the physical forward model of single-pixel imaging.\par     

Subsequently, to impose the sparsity constraint, the intermediate output $O_F$ is processed by the proximal mapping module (described in \secref{sec:proximal_mapping_module}{2.3}). It is first fed into a sparsity encoder to extract deep latent features in the sparse domain:
\begin{equation}
O_{SE} = SE(O_F; \theta_e).
\end{equation}
Next, an ELU-based soft-thresholding operator with a learnable threshold $\zeta$ performs nonlinear shrinkage in the latent space:
\begin{equation}
Feat_{sp} = \text{SoftThreshold}(O_{SE}; \zeta).
\end{equation}
Finally, a sparsity decoder maps the sparse latent representation back to the image domain to generate the refined result:
\begin{equation}
O_P = SD(Feat_{sp}; \theta_d).
\end{equation}
This cascaded encoder-threshold-decoder sequence corresponds to the proximal operator in ISTA. The main procedure is illustrated in Algorithm~\ref{alg:sista_net}. By unifying physics-based optimization with self-supervised deep learning, the proposed SISTA-Net ensures mathematical transparency, thereby enhancing the physical interpretability of SPI reconstruction over neural networks.\par

\vspace*{2mm}\noindent\emph{2.2\quad Fidelity Module}\label{sec:fidelity_module}\vspace*{2mm}

\noindent The fidelity module, instantiated as a custom-designed Res2MM-Net architecture, maps the initial random noise $I$ to an intermediate image representation $O_F$, preserving consistency with physical measurements. To capture both fine-grained local details and global long-range dependencies, we tailor this Res2MM-Net with a hybrid architecture integrating CNNs\cite{he2016deep} and VSSMs\cite{zhu2024vision}, where the latter efficiently models global contexts with linear computational complexity.\par

Specifically, as shown in Fig.~\ref{fig:Net1}(a), the input $I$ is first processed by a head convolution block to expand its channel dimension into a high-dimensional feature space. This head block contains two $3 \times 3$ convolutional layers. To overcome the limited receptive field of standard CNNs, the features are subsequently fed into a sequence of $N$ MamCNN Blocks (described below). By combining local convolutions with Mamba scanning, this sequence acts as a robust measurement consistency fitter. The output of this sequence is added to the head block's output via a skip connection. Finally, a tail block consisting of two $3 \times 3$ convolutional layers refines the features, completing the data-fidelity reconstruction of Res2MM-Net and preparing robust representations for the subsequent proximal mapping.\par

\noindent \textbf{MamCNN Block:} To better capture local and global features while respecting the Deep Image Prior preference for local convolutional extraction\cite{jo2021rethinking}, the MamCNN Block employs a hybrid design of residual convolutions and a Res2Mam Multi-Scale Block (Res2MMB), as shown in Fig.~\ref{fig:Net1}(b). The input features first pass through a $3 \times 3$ convolutional layer before entering the Res2MMB. The resulting features are added to the initial input via a skip connection. Subsequently, this fused representation is fed into a two-layer residual convolutional network for further feature integration.\par

\afterpage{%
\begin{algorithm}[!t]
\footnotesize
\algrenewcommand\algorithmicrequire{\textbf{Input:}}
\algrenewcommand\algorithmicensure{\textbf{Output:}}
\caption{SISTA-Net}
\label{alg:sista_net}
\begin{algorithmic}[1]
\Require
    \Statex Measurements vector $Y \in \R^{N_M}$;
    \Statex Measurement matrix $M \in \R^{N_M \times N_P}$;
    \Statex Hyperparameters $\lambda_1, \lambda_2, \lambda_3$;
    \Statex Max iterations $K_{max}$.
\Ensure
    \Statex Reconstructed Image $x_{final}$.

\State \textbf{Initialization}
\State Randomly initialize $\theta_f,\theta_e,\theta_d$; set $\zeta=0.01$.
\State $I \sim \mathcal{U}(0, 1)^{H \times W}$

\For{$k = 1$ to $K_{max}$}
    \State \textbf{Forward Propagation}
    \State $O_F \leftarrow F(I; \theta_f)$
    \State $O_{SE} \leftarrow SE(O_F; \theta_e)$
    \State $Feat_{sp} \leftarrow \operatorname{sgn}(O_{SE}) \odot \operatorname{ELU}(|O_{SE}|-\zeta)$
    \State $O_P \leftarrow SD(Feat_{sp}; \theta_d)$

    \State \textbf{Loss Calculation}
    \State $\mathcal{L}_{Fidelity} \leftarrow \| M \cdot O_F - Y \|_2^2$
    \State $\mathcal{L}_{Sparsity} \leftarrow \| Feat_{sp} \|_1$
    \State $\mathcal{L}_{Proximal} \leftarrow D_{KL}( M \cdot O_P \| Y )$
    \State $\mathcal{L}_{Total} \leftarrow \lambda_1 \mathcal{L}_{Fidelity} + \lambda_2 \mathcal{L}_{Sparsity} + \lambda_3 \mathcal{L}_{Proximal}$

    \State \textbf{Parameter Update}
    \State $\Theta = \{\theta_f, \theta_e, \theta_d, \zeta\}$
    \State $\Theta \leftarrow \text{Adam}(\nabla_{\Theta}, \mathcal{L}_{Total}, \eta, \beta_1, \beta_2)$
\EndFor

\State \Return $x_{final} \leftarrow O_P$
\end{algorithmic}
\end{algorithm}
}

\noindent \textbf{Res2Mam Multi-Scale Block:} In single-pixel imaging, image resolution directly affects
the complexity of the inverse problem\cite{jia2025100}. Fusing multi-scale features is
crucial for capturing rich context in high-resolution
reconstructions\cite{cho2021rethinking}. Inspired by the multi-scale receptive fields of
Res2Net\cite{gao2019res2net}, we design Res2MMB with hierarchical interactive
capabilities. As shown in Fig.~\ref{fig:Net1}(c), unlike traditional channel-wise partitioning, we
construct three spatial branches with varying downsampling rates to
extract full-scale ($1\times$), half-scale ($1/2\times$), and
quarter-scale ($1/4\times$) features.\par
Given an input $X$, the initial multi-scale extraction is formulated
as:
\begin{equation}
X_1 = \text{Conv}_1(X),X_2 = \text{Conv}_2(X),X_3 = \text{Conv}_4(X),
\end{equation}
where $\text{Conv}_s$ denotes convolutional layers with an
equivalent stride of $s$. To achieve coarse-to-fine global guidance,
long-range features from smaller scales are progressively upsampled
and injected into adjacent larger scales. Each branch then undergoes
efficient state-space scanning via the Window Mamba Block (WMB). This hierarchical
interaction is expressed as:
\begin{equation}
\begin{aligned}
X'_1 &= \text{WMB}_1(X_1 + \text{Up}(X'_2)),\\
X'_2 &= \text{WMB}_2(X_2 + \text{Up}(X'_3)),\\
X'_3 &= \text{WMB}_3(X_3),
\end{aligned}
\end{equation}
where $\text{Up}(\cdot)$ denotes bilinear upsampling followed by
convolutional smoothing. Finally, all branch features are restored to their
original spatial resolution, concatenated along the channel
dimension, and fused via a $1 \times 1$ convolution:
\begin{equation}
\begin{aligned}
\hat{X} &= \text{Concat}\big(C_1(X'_1), C_2(\text{Up}(X'_2)), C_3(\text{Up}(X'_3))\big),\\
Y &= \text{Conv}_{1\times1}(\hat{X}),
\end{aligned}
\end{equation}
where $C_i$ represents the alignment convolution at the end of each branch. By combining multi-scale spatial features, the Res2MMB significantly enhances the representation capacity for reconstructions.\par

\noindent \textbf{Window Mamba Block:} Traditional VSSMs directly flatten 2D feature maps into 1D sequences for state-space iteration \cite{yu2025mambaout}, which severely disrupts the local 2D spatial structures that the DIP framework fundamentally relies on as an inductive bias \cite{cheng2019bayesian}. To resolve this conflict, WMB adopts a cascaded strategy: local window modeling followed by global long-range scanning. Specifically, the input $X_{\text{in}} \in \mathbb{R}^{B \times C \times H \times W}$ is first partitioned into non-overlapping local windows via a spatial partitioning operation $\mathcal{P}(\cdot)$, as illustrated in Fig.~\ref{fig:Net1}(d). A Vision State Space Block (VSSB) is applied within each independent window to capture fine-grained local structures. The features are then restored to their original dimensions using a reverse operation $\mathcal{R}(\cdot)$:
\begin{equation}
X_{\text{local}} = \mathcal{R}(\text{VSSB}(\mathcal{P}(X_{\text{in}}))).
\end{equation}
After capturing local correlations through two successive fine-grained state-space scans, the features are forwarded to a global VSSB for full-image interaction:
\begin{equation}
X_{\text{out}} = \text{VSSB}(X_{\text{local}}).
\end{equation}
By unifying local and global scanning, WMB preserves the 2D spatial prior and aligns perfectly with the locality preference of DIP, achieving an optimal balance between local detail retention and global structural consistency.\par

\noindent \textbf{Vision State Space Block:} VSSB is used for global context modeling. As shown in Fig.~\ref{fig:Net1}(e), a $3 \times 3$ depthwise convolution first enhances local context, followed by 2D Selective Scan (SS2D) along multiple spatial directions\cite{liu2024vmamba}. A parallel gating branch adaptively weights feature channels and suppresses irrelevant responses\cite{shi2018lowLight}, yielding non-local feature interaction while preserving useful local cues.\par

\vspace*{2mm}\noindent\emph{2.3\quad Proximal Mapping
Module}\label{sec:proximal_mapping_module}\vspace*{2mm}

\noindent The proximal mapping module replaces the fixed sparse transform in ISTA with a learnable encoder-threshold-decoder structure. Given $O_F$, the Sparsity Encoder (SE) maps it into a latent sparse domain, the Learnable Soft-Thresholding (LST) operator removes weak coefficients, and the Sparsity Decoder (SD) maps the filtered features back to the image domain. This design provides an adaptive proximal operator for self-supervised SPI reconstruction.\par

\begin{figure}[htbp]
\centering
\includegraphics[width=\columnwidth]{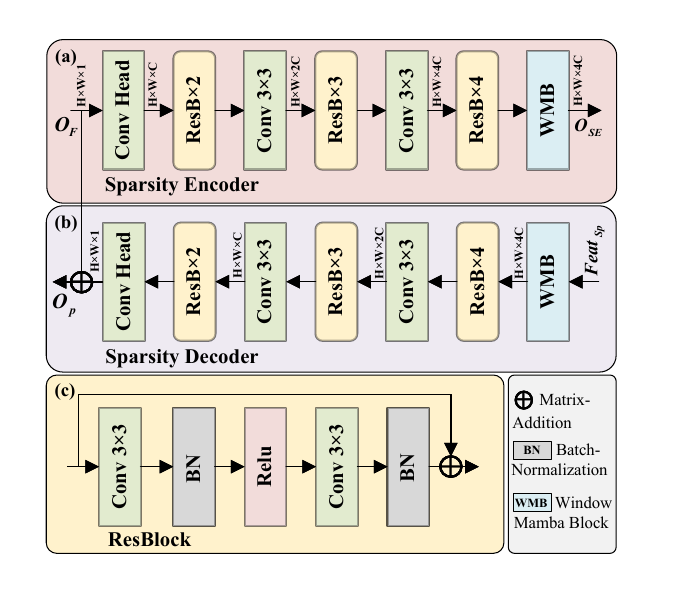}
\caption{Architecture of the Proximal Mapping Module. (a) The structure of the Sparsity Encoder. (b) The structure of the Sparsity Decoder. (c) The structure of the ResBlock.}
\label{fig:Net2}
\end{figure}

\noindent \textbf{Sparsity Encoder:} The SE learns a nonlinear sparse transform $\mathcal{F}(\cdot)$ by combining convolutional projection, residual feature refinement, and a WMB for long-range dependency modeling, as illustrated in Fig.~\ref{fig:Net2}(a). Its forward process is formulated as:
\begin{equation}
O_{\text{SE}} = \text{WMB}_{\text{en}}\Big( \text{ResSeq}\big(
\text{Trans}(O_F) \big) \Big).
\end{equation}
The resulting feature $O_{\text{SE}}$ provides a compact latent representation for subsequent thresholding.\par

\noindent \textbf{Learnable Soft-Thresholding:} To avoid manual threshold tuning, we adopt an ELU-based learnable soft-thresholding operator with a scalar threshold $\zeta$, initialized to 0.01 and jointly optimized with the network parameters. Given the latent feature $O_{\text{SE}}$, the thresholding operation is defined as:
\begin{equation}
\text{Feat}_{\text{sp}} = \operatorname{sgn}(O_{\text{SE}}) \odot \operatorname{ELU}\left(|O_{\text{SE}}|-\zeta\right),
\end{equation}
where $\operatorname{ELU}(u)=u$ for $u>0$ and $\operatorname{ELU}(u)=\exp(u)-1$ otherwise. For coefficients above the learned threshold, the operator performs magnitude shrinkage, while the smooth negative branch of ELU maintains gradient propagation for responses below the threshold. Together with the $L_1$ sparsity loss, this threshold-controlled transformation promotes concentration of the latent responses and improves robustness to noise.\par

\noindent \textbf{Sparsity Decoder:} The SD acts as the inverse transform $\tilde{\mathcal{F}}(\cdot)$, restoring the thresholded latent feature $\text{Feat}_{\text{sp}}$ to the image domain. As shown in Fig.~\ref{fig:Net2}(b), it mirrors the encoder with a decoding WMB, residual refinement, transition blocks, and a tail convolution:
\begin{equation}
\begin{aligned}
Z_{\text{de}} &= \text{ResSeq}\big(\text{WMB}_{\text{de}}(\text{Feat}_{\text{sp}})\big),\\
O_P &= \text{TailConv}\big(\text{Trans}(Z_{\text{de}})\big).
\end{aligned}
\end{equation}

\noindent These components form a learnable proximal sparse solver that filters redundant features while preserving reconstruction structure.\par

\vspace*{2mm}\noindent\emph{2.4\quad Loss Functions}\vspace*{2mm}

\noindent Operating in a self-supervised manner, our network relies on the measurements $Y$ to compute a total loss function comprising three complementary components.\par
\noindent\textbf{Fidelity Loss} ($\mathcal{L}_{\text{Fidelity}}$): To ensure the intermediate image $O_F$ aligns with the physical measurement process, we compute the Mean Squared Error (MSE) between the predicted intensities and the actual measurements:
\begin{equation}
\mathcal{L}_{\text{Fidelity}} = \frac{1}{N} \| M O_F - Y \|_2^2.
\end{equation}
\textbf{Sparsity Loss} ($\mathcal{L}_{\text{Sparsity}}$): Inspired by compressive sensing theory, we apply an $L_1$-norm constraint to the soft-thresholded features, which enforces physical sparsity in the latent feature domain:
\begin{equation}
\mathcal{L}_{\text{Sparsity}} = \frac{1}{N} \| \text{Feat}_{\text{sp}} \|_1.
\end{equation}

\noindent\textbf{Proximal Response Loss} ($\mathcal{L}_{\text{Proximal}}$): To stabilize the decoded proximal output, we constrain its measurement response using KL divergence. Softmax normalization converts the measured and predicted intensities into positive relative-response profiles, with $P_i=\exp(Y_i)/\sum_j\exp(Y_j)$ and $Q_i=\exp((MO_P)_i)/\sum_j\exp((MO_P)_j)$. Absolute measurement fidelity is already enforced by the Fidelity Loss on $MO_F$. Applying another MSE term to $MO_P$ would emphasize pointwise deviations caused by noise and interference, potentially weakening the noise-suppression capability of soft-thresholding. Therefore, KL divergence is used to preserve the relative measurement-response structure after proximal mapping without reinforcing suppressed noise amplitudes. The proximal loss is:

\begin{equation}
\mathcal{L}_{\text{Proximal}} = D_{\text{KL}}(P \| Q) = \sum P \log\left(\frac{P}{Q}\right).
\end{equation}

\noindent Finally, the network parameters are updated by minimizing the weighted total loss:
\begin{equation}
\mathcal{L}_{\text{Total}} = \lambda_1 \mathcal{L}_{\text{Fidelity}} + \lambda_2 \mathcal{L}_{\text{Sparsity}} + \lambda_3\mathcal{L}_{\text{Proximal}},
\end{equation}
where $\lambda_1$, $\lambda_2$, and $\lambda_3$ are hyperparameters balancing these loss terms.\par

\vspace*{2mm}\noindent\emph{2.5\quad Optimization Alignment with Traditional ISTA}\label{sec:ista_alignment}\vspace*{2mm}

\noindent To clarify the ISTA-inspired optimization behavior of SISTA-Net, we analyze how the self-supervised parameter update induces an image-space descent. In this view, the fidelity network provides the intermediate reconstruction, while the encoder-threshold-decoder module acts as a learnable proximal operator.\par

\noindent Let the intermediate image and proximal output at the $k$-th iteration be
\begin{equation}
\begin{aligned}
\mathbf{x}^{(k)} &= O_F^{(k)} = F(I;\theta_f^{(k)}),\\
\mathbf{p}^{(k)} &= O_P^{(k)} = G(\mathbf{x}^{(k)};\theta_p^{(k)}),
\end{aligned}
\end{equation}
where $G(\cdot;\theta_p)$ denotes the whole proximal module $SD(T_{\zeta}(SE(\cdot;\theta_e));\theta_d)$ and $\theta_p=\{\theta_e,\theta_d,\zeta\}$. With $\mathbf{z}(\mathbf{x})=T_{\zeta}(SE(\mathbf{x};\theta_e))$, $P=\text{Softmax}(Y)$, and $Q(\mathbf{x})=\text{Softmax}(MG(\mathbf{x};\theta_p))$, the self-supervised objective can be written as a composite energy over the image variable:
\begin{equation}
\begin{aligned}
\mathcal{E}(\mathbf{x})
&= \frac{\lambda_1}{N}\|M\mathbf{x}-Y\|_2^2
+ \frac{\lambda_2}{N}\|\mathbf{z}(\mathbf{x})\|_1\\
&\quad + \lambda_3D_{\text{KL}}\left(P\|Q(\mathbf{x})\right),
\end{aligned}
\end{equation}
where the three terms are measurement fidelity, latent sparsity, and post-proximal measurement consistency.\par

\noindent The image-space gradient of this energy is decomposed as
\begin{equation}
\begin{aligned}
\nabla_{\mathbf{x}}\mathcal{E}(\mathbf{x})
&= \mathbf{g}_{F} + \mathbf{g}_{S} + \mathbf{g}_{P}\\
&= \frac{2\lambda_1}{N}M^{\top}(M\mathbf{x}-Y)
 + \frac{\lambda_2}{N}J_{\mathbf{z}}^{\top}\mathbf{s}\\
&\quad + \lambda_3J_G^{\top}M^{\top}(Q-P),
\end{aligned}
\label{eq:composite_gradient}
\end{equation}
where $\mathbf{g}_{F}$, $\mathbf{g}_{S}$, and $\mathbf{g}_{P}$ denote the fidelity, sparsity, and proximal-response gradients, respectively. Here, $J_{\mathbf{z}}=\partial\mathbf{z}/\partial\mathbf{x}$, $J_G=\partial G/\partial\mathbf{x}$, and $\mathbf{s}\in\partial\|\mathbf{z}(\mathbf{x})\|_1$. Since $\nabla_{\mathbf{u}}D_{\text{KL}}(P\|\text{Softmax}(\mathbf{u}))=\text{Softmax}(\mathbf{u})-P$ with $\mathbf{u}=MG(\mathbf{x};\theta_p)$, the last term becomes a post-proximal back-projected residual.\par

\noindent Let $J_f^{(k)}=\partial F(I;\theta_f^{(k)})/\partial\theta_f$. A gradient update of $\theta_f$ gives
\begin{equation}
\theta_f^{(k+1)} = \theta_f^{(k)} - \eta\left(J_f^{(k)}\right)^{\top}\nabla_{\mathbf{x}}\mathcal{E}(\mathbf{x}^{(k)}).
\end{equation}
Using a first-order Taylor expansion of the network output,
\begin{equation}
\mathbf{x}^{(k+1)} \approx \mathbf{x}^{(k)} + J_f^{(k)}\left(\theta_f^{(k+1)}-\theta_f^{(k)}\right).
\end{equation}
Substitution gives the implicit image-space update
\begin{equation}
\mathbf{x}^{(k+1)} \approx \mathbf{x}^{(k)} - \eta K_f^{(k)}\nabla_{\mathbf{x}}\mathcal{E}(\mathbf{x}^{(k)}),
\label{eq:image_space_update}
\end{equation}
where $K_f^{(k)}=J_f^{(k)}(J_f^{(k)})^{\top}$ is a positive semi-definite neural preconditioner. With Adam, this preconditioner becomes $J_f^{(k)}A^{(k)}(J_f^{(k)})^{\top}$, where $A^{(k)}$ is the positive adaptive scaling matrix.\par

\noindent Substituting Eq.~\eqref{eq:composite_gradient} into Eq.~\eqref{eq:image_space_update} yields
\begin{equation}
\mathbf{x}^{(k+1)} \approx \mathbf{x}^{(k)} - \eta K_f^{(k)}\left(\mathbf{g}_{F}^{(k)}+\mathbf{g}_{S}^{(k)}+\mathbf{g}_{P}^{(k)}\right).
\label{eq:sista_full_update}
\end{equation}
The first term has the same measurement-residual back-projection form as the classical ISTA data step:
\begin{equation}
\mathbf{r}^{(k)}
=
\mathbf{x}^{(k)}
-
\rho M^{\top}(M\mathbf{x}^{(k)}-Y),
\label{eq:ista_gradient_step}
\end{equation}
with the scalar step size replaced by an adaptive matrix step,
\begin{equation}
\rho^{(k)}_{\text{eff}}
=
\eta\frac{2\lambda_1}{N}K_f^{(k)}.
\end{equation}
Thus, the self-supervised update of SISTA-Net can be interpreted as a ISTA-inspired image-space descent, where the fidelity module provides a learned data-consistency update and the proximal module imposes sparse regularization with measurement consistency. With $\lambda_1=10$, $\lambda_2=0.1$, and $\lambda_3=1$, this induced image-space descent is primarily governed by measurement fidelity.\par

\vspace*{2mm} \noindent \textbf{3\quad Simulation Experiments}\vspace*{2mm}

\noindent To verify the effectiveness of the proposed SISTA-Net, we conducted comparative experiments on simulated datasets with $320 \times 320$ binary and grayscale natural images as reconstruction targets. Peak Signal-to-Noise Ratio (PSNR), Structural Similarity (SSIM), and high-frequency PSNR (PSNR-HF) were used to evaluate pixel-level fidelity, structural preservation, and high-frequency detail fidelity, respectively, with higher values indicating better reconstruction quality.

\begin{figure}[!b]
\centering
\includegraphics[width=0.98\columnwidth]{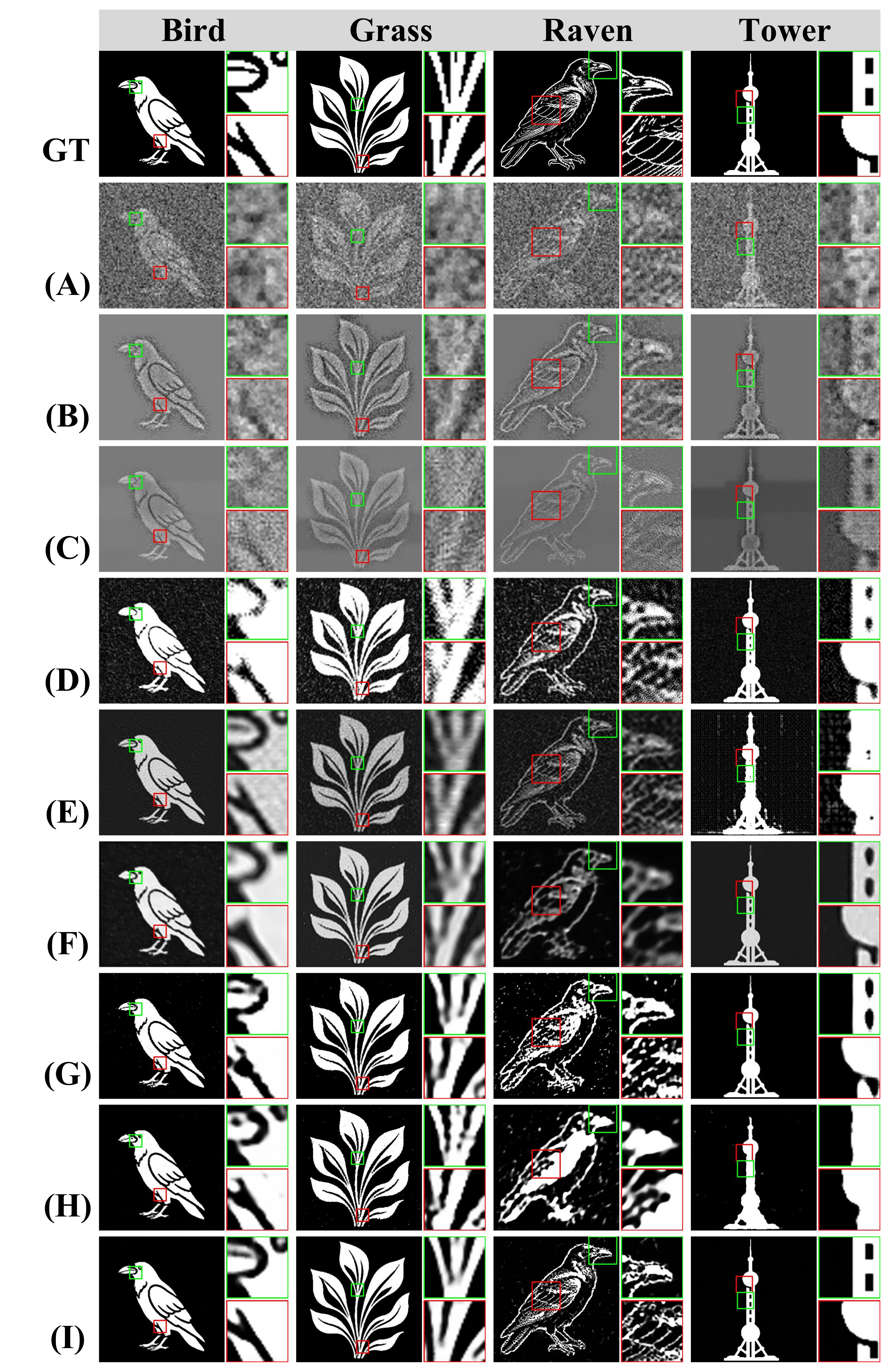}
\caption{Reconstruction results on binary images. The markers (A)--(I) denote DGI, ISTA-Net++, TransCS, Restormer, URNet, GILM, UNet, GIDC, and Ours in order.}
\label{fig:ex1}
\end{figure}

\begin{table*}[!htbp]
\centering
\small
\caption{Quantitative comparison for simulated binary and grayscale reconstructions.}
\label{tab:simulated-quality}
\resizebox{\textwidth}{!}{%
\begin{tabular}{lcccccccccccc}
\toprule
\multirow{2}{*}{Method} & \multicolumn{3}{c}{Bird} & \multicolumn{3}{c}{Grass} & \multicolumn{3}{c}{Raven} & \multicolumn{3}{c}{Tower} \\
\cmidrule(lr){2-4}\cmidrule(lr){5-7}\cmidrule(lr){8-10}\cmidrule(lr){11-13}
    & PSNR & SSIM & PSNR-HF & PSNR & SSIM & PSNR-HF & PSNR & SSIM & PSNR-HF & PSNR & SSIM & PSNR-HF \\
\midrule
\textbf{DGI} & 7.67 & 0.0288 & 18.40 & 6.91 & 0.0442 & 15.79 & 7.30 & 0.0224 & 13.00 & 7.38 & 0.0213 & 20.37 \\
\textbf{ISTA-Net++} & 6.23 & 0.0497 & 19.52 & 7.19 & 0.0875 & 16.27 & 7.08 & 0.0390 & 13.40 & 7.16 & 0.0273 & 22.15 \\
\textbf{TransCS} & 6.47 & 0.0400 & 19.36 & 6.79 & 0.0542 & 15.93 & 7.02 & 0.0212 & 13.19 & 9.26 & 0.0257 & 22.18 \\
\textbf{Restormer} & 14.18 & 0.1375 & 16.21 & 12.40 & 0.2414 & 14.47 & 10.45 & 0.0589 & 11.68 & 17.76 & 0.0823 & 19.58 \\
\textbf{URNet} & 16.44 & 0.1297 & 21.36 & 13.10 & 0.1775 & 17.07 & 12.48 & 0.0653 & 13.05 & 14.14 & 0.0512 & 18.59 \\
\textbf{GILM} & 17.63 & 0.1565 & 20.51 & 14.74 & 0.2660 & 17.88 & 12.30 & 0.1760 & 13.25 & 18.32 & 0.0961 & 25.50 \\
\textbf{UNet} & 23.24 & 0.9278 & 23.72 & 20.37 & 0.9288 & 20.90 & 10.89 & 0.5570 & 12.48 & 27.71 & 0.9443 & 27.79 \\
\textbf{GIDC} & 23.48 & 0.9499 & 23.46 & 20.87 & 0.9247 & 21.11 & 9.21 & 0.4937 & 13.09 & 18.37 & 0.9143 & 23.21 \\
\textbf{Ours} & \textbf{24.64} & \textbf{0.9588} & \textbf{24.64} & \textbf{23.01} & \textbf{0.9596} & \textbf{23.02} & \textbf{13.22} & \textbf{0.5990} & \textbf{13.42} & \textbf{32.81} & \textbf{0.9873} & \textbf{32.85} \\
\bottomrule
\end{tabular}
}

\vspace{1mm}
\resizebox{\textwidth}{!}{%
\begin{tabular}{lcccccccccccc}
\toprule
\multirow{2}{*}{Method} & \multicolumn{3}{c}{Clown} & \multicolumn{3}{c}{Crab} & \multicolumn{3}{c}{Fish} & \multicolumn{3}{c}{Octopus} \\
\cmidrule(lr){2-4}\cmidrule(lr){5-7}\cmidrule(lr){8-10}\cmidrule(lr){11-13}
    & PSNR & SSIM & PSNR-HF & PSNR & SSIM & PSNR-HF & PSNR & SSIM & PSNR-HF & PSNR & SSIM & PSNR-HF \\
\midrule
\textbf{DGI} & 12.63 & 0.1396 & 21.42 & 13.25 & 0.1330 & 20.51 & 12.69 & 0.1435 & 21.44 & 12.29 & 0.1291 & 20.03 \\
\textbf{ISTA-Net++} & 13.41 & 0.4496 & 23.94 & 14.60 & 0.4745 & 22.75 & 15.86 & 0.5188 & 23.77 & 13.44 & 0.4255 & 22.30 \\
\textbf{TransCS} & 12.80 & 0.2834 & 22.38 & 15.76 & 0.3439 & 21.52 & 14.80 & 0.3157 & 21.98 & 13.74 & 0.2539 & 20.14 \\
\textbf{Restormer} & 21.30 & 0.4649 & 24.00 & 17.60 & 0.2795 & 20.75 & 19.05 & 0.4132 & 23.16 & 17.29 & 0.3533 & 20.99 \\
\textbf{URNet} & 21.73 & 0.5493 & 25.69 & 18.73 & 0.2894 & 22.57 & 20.35 & 0.4681 & 24.20 & 19.23 & 0.4862 & 22.54 \\
\textbf{GILM} & 20.36 & 0.6052 & 25.30 & 17.56 & 0.3425 & 21.54 & 19.76 & 0.4674 & 23.07 & 17.43 & 0.2951 & 22.45 \\
\textbf{UNet} & 21.56 & 0.5508 & 24.48 & 15.31 & 0.3379 & 20.31 & 18.89 & 0.5017 & 24.43 & 17.95 & 0.4713 & 20.24 \\
\textbf{GIDC} & 22.22 & 0.6260 & 25.20 & 15.85 & 0.4148 & 21.31 & 16.63 & 0.4956 & 22.76 & 17.32 & 0.5037 & 21.89 \\
\textbf{Ours} & \textbf{22.67} & \textbf{0.6302} & \textbf{25.95} & \textbf{19.21} & \textbf{0.4993} & \textbf{23.03} & \textbf{20.85} & \textbf{0.5651} & \textbf{24.57} & \textbf{19.52} & \textbf{0.5551} & \textbf{22.78} \\
\bottomrule
\end{tabular}
}
\end{table*}

\begin{figure}[!h]
\centering
\includegraphics[width=0.98\columnwidth]{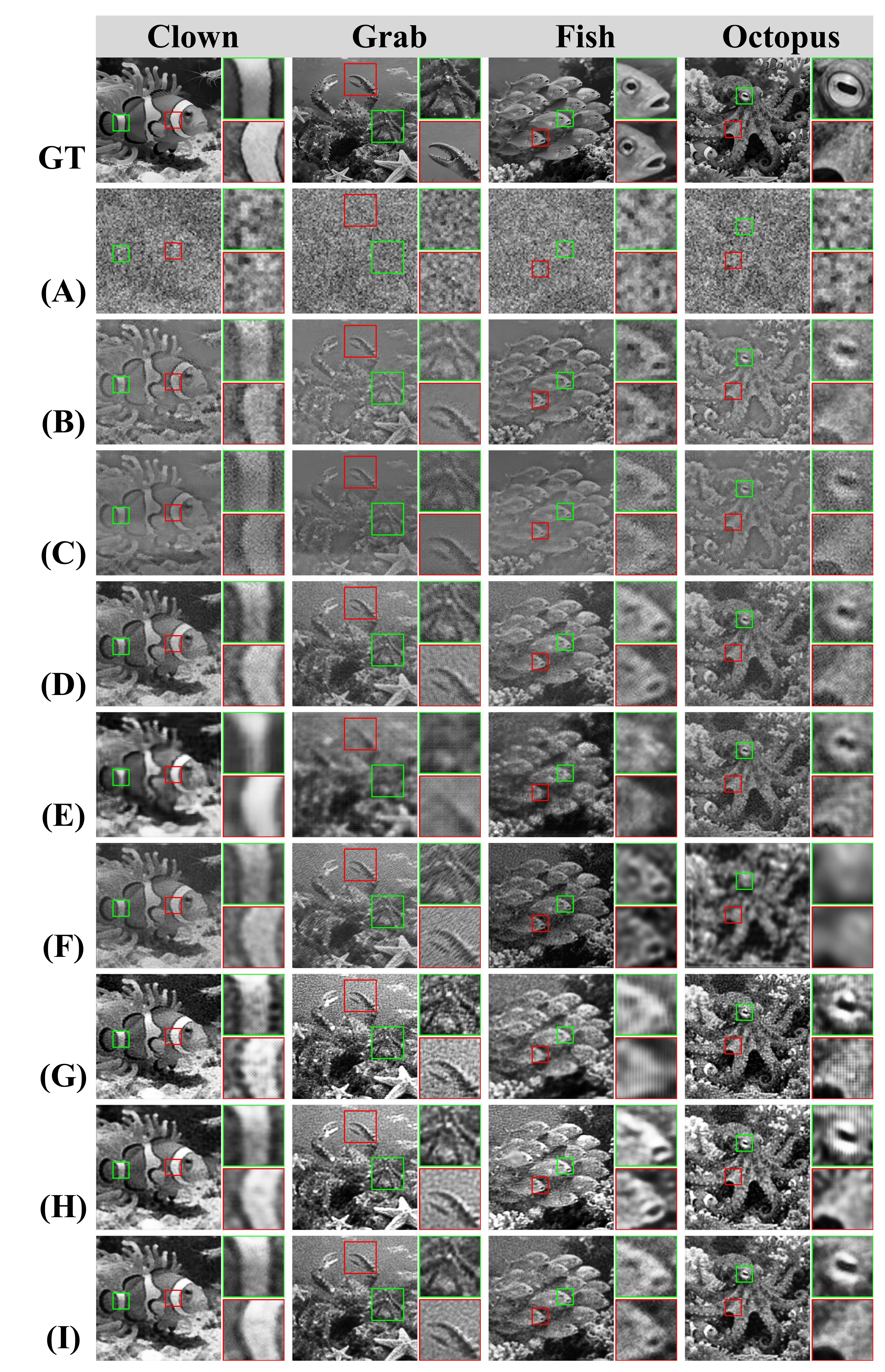}
\caption{Reconstruction results on complex grayscale natural images.}
\label{fig:ex2}
\end{figure}

\begin{figure*}[!htbp]
\centering
\includegraphics[width=0.93\textwidth,keepaspectratio]{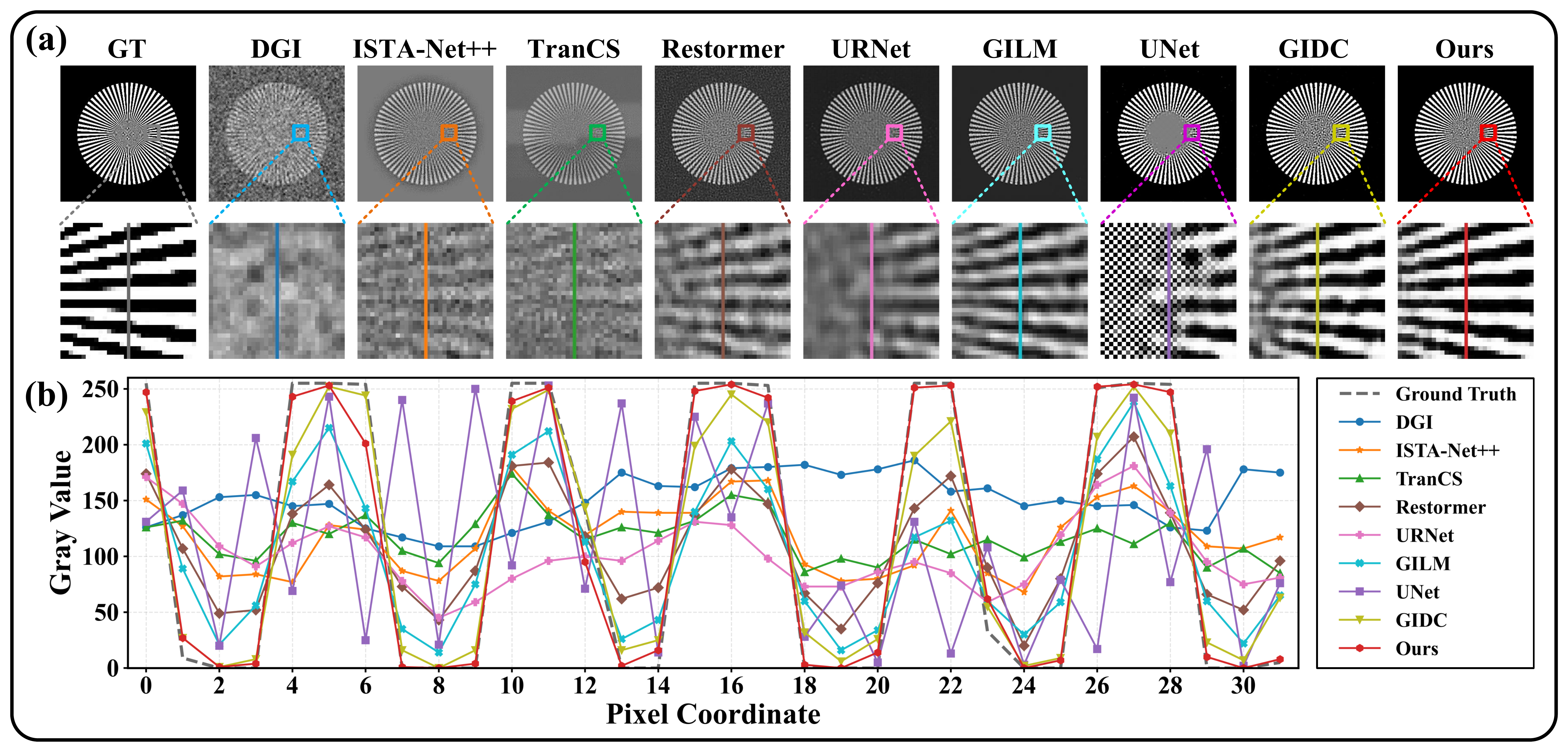}
\caption{Detail reconstruction results. (a) Reconstructed images and their locally enlarged regions. (b) Vertical grayscale intensity profiles at the mark.}
\label{fig:ex3}
\end{figure*}

The compared methods cover representative physics-based, deep unfolding, general restoration, and self-supervised reconstruction paradigms. DGI\cite{ferri2010differential} was selected as the traditional correlation baseline. ISTA-Net++\cite{you2021ista} and TransCS\cite{shen2022transcs} were selected as deep unfolding networks for compressive sensing. UNet\cite{ronneberger2015u} was used as a generic encoder-decoder network, and Restormer\cite{zamir2022restormer} was adopted as an attention-based general image restoration network. URNet\cite{li2023urnet}, GILM\cite{chen2025large}, and GIDC\cite{wang2022far} were included as reconstruction networks used under the self-supervised measurement-consistency paradigm. All deep learning models utilized uniformly distributed random noise as input and 1D measurement signals as the sole supervisory signal, with 2,000 training iterations. Experiments were implemented using the PyTorch framework on an NVIDIA A100 GPU with 40 GB of VRAM. The initial learning rate was set to 0.001 for UNet and 0.0001 for the other deep models, decaying to 0.00001 via a cosine annealing schedule. We adopt PSNR-based early stopping with a patience of 20 for all simulation tests, where ground truth is available.This strategy mitigates late-stage overfitting and provides a consistent checkpoint selection criterion for fair comparison among different methods.\par
\par

\vspace*{2mm}\noindent\emph{3.1\quad Reconstruction Quality for Simulated
Objects}\label{sec:image_quality_simulated}\vspace*{2mm}

\noindent In this experiment, we reconstructed binary and grayscale images using 3,000 and 6,000 measurements with sampling ratios of 2.93\% and 5.86\%, respectively.\par

Fig.~\ref{fig:ex1} illustrates the binary reconstruction results. DGI produced severe background noise and structural loss because correlation-based recovery is unstable under such a low sampling ratio. ISTA-Net++ and TransCS also yielded weak-contrast reconstructions after being adapted to the self-supervised SPI setting, since their supervised block-wise priors cannot be fully activated without paired training data. Restormer, URNet, GILM, UNet, and GIDC recovered more recognizable contours, but still suffered from residual artifacts, noise accumulation, or blurred details. In contrast, SISTA-Net achieved the best visual and quantitative performance by combining measurement fidelity with proximal sparsity, effectively suppressing dense noise while preserving target structures.\par

For grayscale scenes in Fig.~\ref{fig:ex2}, the overall trend is consistent but the richer textures make reconstruction more challenging. Baseline methods either lost fine structures or introduced noticeable noise, whereas SISTA-Net retained clearer boundaries and texture details, reflecting a better balance between detail recovery and noise suppression. The quantitative results in Table~\ref{tab:simulated-quality} further confirm the visual comparison.\par

\vspace*{2mm}\noindent\emph{3.2\quad Detail Recovery for Simulated
Object}\vspace*{2mm}

\noindent Accurate detail recovery is a core requirement of SPI, as detail preservation directly determines its utility in practical applications like industrial inspection and remote sensing. To quantitatively evaluate detail reconstruction, we selected a target image featuring dense fine stripes with a resolution of $320\times 320$. We acquired 10,000 measurements, corresponding to a 9.77\% sampling ratio, keeping all other settings consistent with \secref{sec:image_quality_simulated}{3.1}.\par

\begin{figure}[!b]
\centering
\includegraphics[width=0.99\columnwidth,keepaspectratio]{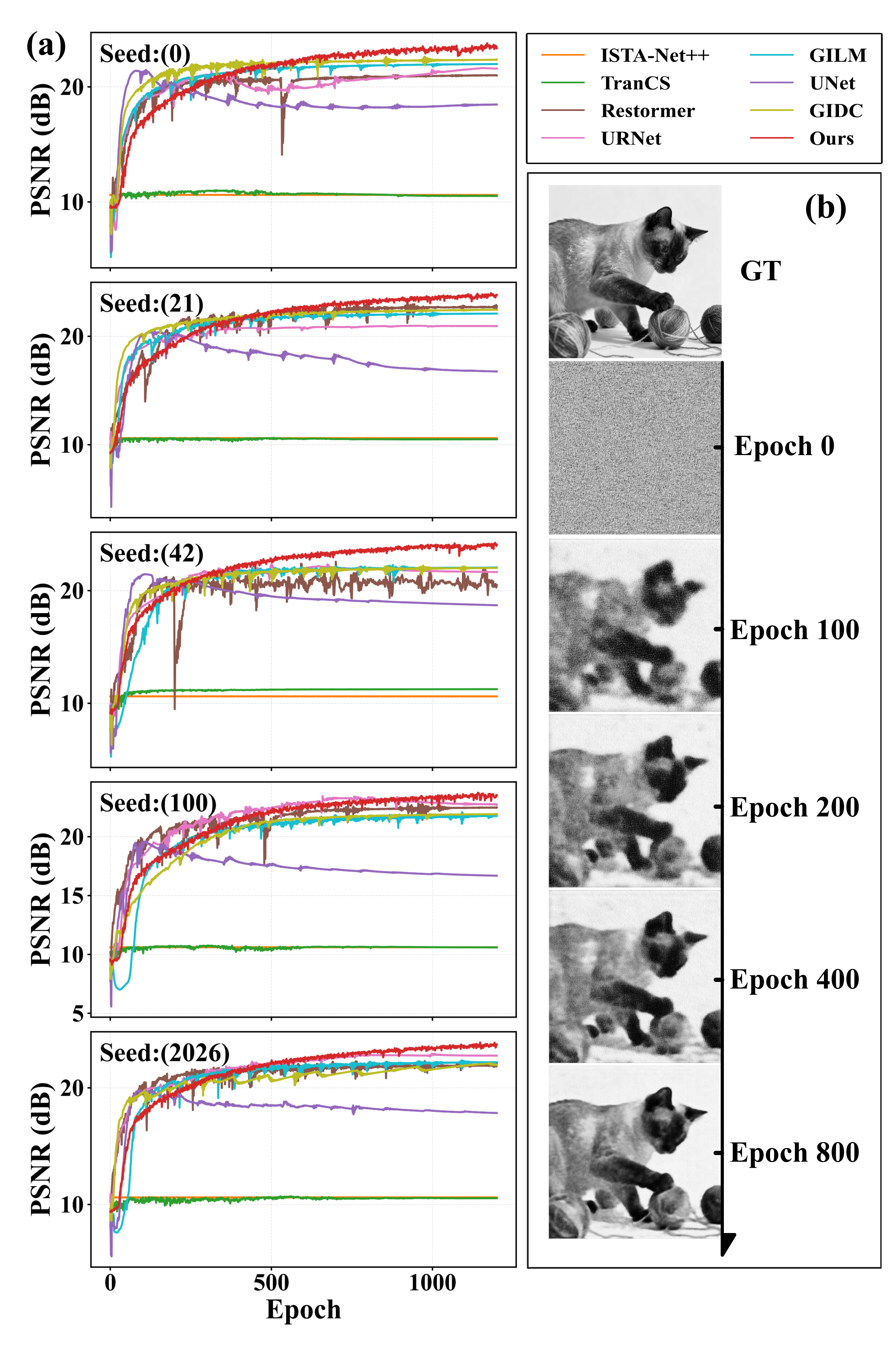}
\caption{Convergence and stability analysis. (a) PSNR curves of different models under five random seeds. (b) Intermediate SISTA-Net reconstruction results at selected epochs under random seed 0.}
\label{fig:efficiency}
\end{figure}

\begin{figure*}[!t]
\centering
\includegraphics[width=0.88\textwidth,keepaspectratio]{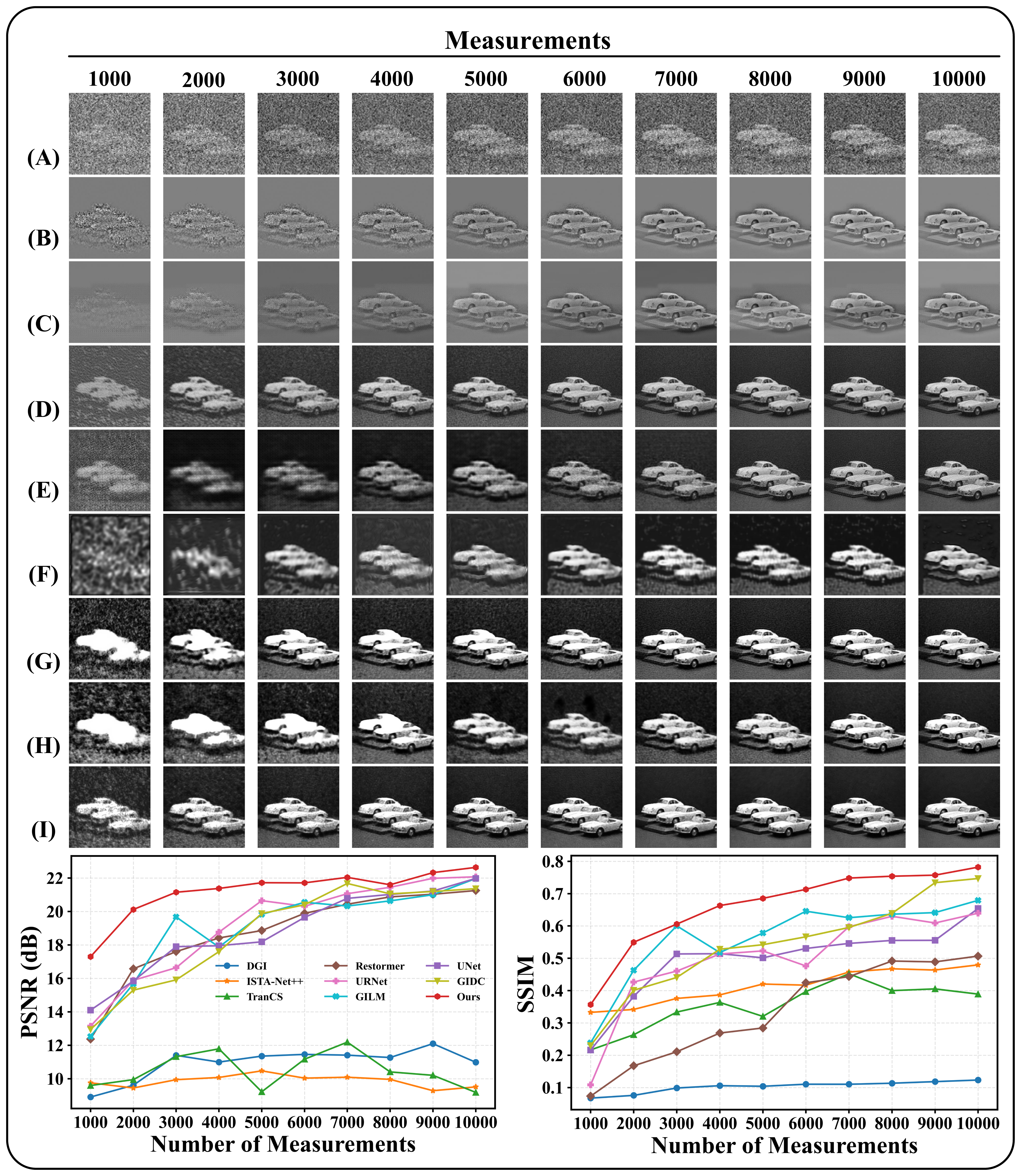}
\caption{Reconstruction results over different sampling ratios. The markers (A)--(I) denote DGI, ISTA-Net++, TransCS, Restormer, URNet, GILM, UNet, GIDC, and Ours in order.}
\label{fig:ex4}
\end{figure*}

As can be seen from Fig.~\ref{fig:ex3}(a), the locally magnified results clearly demonstrate SISTA-Net's distinct advantage in detail restoration. Most baseline methods suffer from severe signal-noise crosstalk, resulting in fused structures and boundary aliasing within dense stripe regions. The vertical grayscale profiles in Fig.~\ref{fig:ex3}(b) further corroborate this observation, as SISTA-Net's curve closely matches the ground truth and exhibits sharp step transitions at the light-dark boundaries. Conversely, the smoothed and fluctuating curves of the baseline methods indicate a weakened contrast that fails to resolve fine structures. These results fully demonstrate that SISTA-Net's hybrid architecture and explicit sparsity constraint effectively preserve high-frequency details.\par

\vspace*{2mm}\noindent\emph{3.3\quad Reconstruction Quality at Different Sample Rates}\vspace*{2mm}

\noindent Achieving high-quality reconstruction under low sampling ratios is essential for SPI to reduce data acquisition time. To evaluate sampling ratio adaptability, we tested ten ratios ranging from 0.97\% to 9.77\% (1,000 to 10,000 measurements) on a $320 \times 320$ target image, keeping all other settings consistent with \secref{sec:image_quality_simulated}{3.1}.\par

As shown in Fig.~\ref{fig:ex4}, SISTA-Net consistently maintained the best reconstruction quality across sampling ratios, with the most evident advantage under extreme undersampling. At 0.97\% and 2.93\%, baseline methods suffered from weak contrast, residual interference, or blurred details, whereas SISTA-Net preserved target structures and cleaner backgrounds. As the sampling ratio increased, all methods improved and the metric gaps gradually narrowed, while SISTA-Net remained the best overall, indicating stronger sampling-ratio robustness.\par

\vspace*{2mm}\noindent\emph{3.4\quad Convergence and Initialization Stability}\label{sec:convergence_stability}\vspace*{2mm}

\noindent To evaluate convergence and initialization stability, we used a $320\times 320$ target and acquired 10,000 measurements, corresponding to a 9.77\% sampling ratio. Five random seeds, namely 0, 21, 42, 100, and 2026, were used to generate different random initializations, and the PSNR evolution was recorded for each model. As shown in Fig.~\ref{fig:efficiency}(a), all deep models improved rapidly in the early stage and then gradually converged. SISTA-Net consistently reached the best late-stage PSNR across five seeds, indicating stable optimization under different random initializations. The intermediate reconstructions in Fig.~\ref{fig:efficiency}(b) further show a progressive transition from noisy patterns to clear target structures, supporting the robustness of its physical-consistency optimization.\par

\begin{figure}[!t]
\centering
\includegraphics[width=0.98\columnwidth,keepaspectratio]{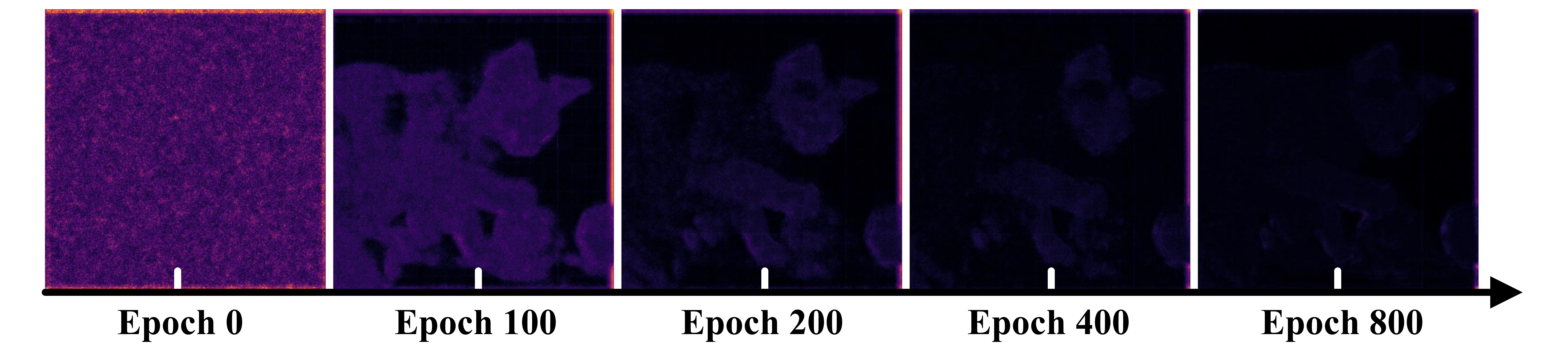}
\caption{Evolution of sparse latent feature maps.}
\label{fig:sparse-feature}
\end{figure}

\vspace{1mm}
\begin{table}[!h]
\centering
\scriptsize
\caption{Latent feature metrics.}
\label{tab:sparse-feature}
\resizebox{\columnwidth}{!}{%
\begin{tabular}{cccccc}
\toprule
SR(N) & Epoch & Mean & $E_{1\%}$ & $E_{3\%}$ & $E_{5\%}$ \\
\midrule
\multirow{3}{*}{2.93\% (3k)} & 200 & 0.0393 & 0.7821 & 0.8717 & 0.9123 \\
    & 400 & 0.0051 & 0.9586 & 0.9812 & 0.9891 \\
    & 800 & 0.0014 & 0.9521 & 0.9796 & 0.9884 \\
\midrule
\multirow{3}{*}{5.86\% (6k)} & 200 & 0.0250 & 0.8648 & 0.9249 & 0.9501 \\
    & 400 & 0.0033 & 0.9697 & 0.9869 & 0.9925 \\
    & 800 & 0.0010 & 0.9652 & 0.9860 & 0.9922 \\
\midrule
\multirow{3}{*}{9.77\% (10k)} & 200 & 0.0163 & 0.9133 & 0.9535 & 0.9697 \\
    & 400 & 0.0010 & 0.9303 & 0.9761 & 0.9871 \\
    & 800 & 0.0006 & 0.9632 & 0.9864 & 0.9923 \\
\bottomrule
\end{tabular}
}%
\end{table}
\vspace{1mm}

To further validate the sparse latent assumption, we visualized the soft-thresholded features corresponding to the intermediate reconstructions in Fig.~\ref{fig:efficiency}(b). As shown in Fig.~\ref{fig:sparse-feature}, diffuse early responses gradually shrink into compact high-energy regions. Table~\ref{tab:sparse-feature} reports the sparse-feature metrics under different sampling settings, where SR(N) denotes the sampling ratio and the number of patterns. The energy ratio $E_{k\%}$ denotes the fraction of total latent energy contained in the largest-magnitude $k\%$ coefficients. Across all settings, optimization reduces the mean latent response and increases $E_{1\%}$, $E_{3\%}$, and $E_{5\%}$, indicating a more concentrated latent representation.\par

\vspace*{2mm}\noindent\emph{3.5\quad Noise Robustness and Adaptive Threshold Analysis}\label{sec:noise_robustness}\vspace*{2mm}

\noindent Following the SPI noise models in \cite{degradations-for-GI}, practical measurements are mainly affected by Gaussian electronic and readout noise and Poisson photon shot noise. We therefore simulated these components separately and jointly on the noise-free bucket signal $s_i$. Gaussian noise followed $d_i^{\mathrm{G}}=s_i+n_{g,i}$ with $n_{g,i}\sim\mathcal{N}(0,\sigma_g^2)$, where $\mathrm{RMS}_s=\sqrt{M^{-1}\sum_i s_i^2}$ and $\sigma_g=\mathrm{RMS}_s10^{-\mathrm{SNR}_{\mathrm{dB}}/20}$. Poisson noise followed $N_i\sim\mathrm{Pois}(Ks_i)$ and $d_i^{\mathrm{P}}=N_i/K$, while mixed noise followed $d_i^{\mathrm{M}}=N_i/K+n_{g,i}$. Lower SNR and smaller $K$ represent stronger Gaussian and photon noise, respectively, where $K$ denotes the photon-count scale. Tests used a 9.77\% sampling ratio, covering SNR from 60 to 40~dB, $K$ from $1\times10^6$ to $1\times10^4$, and five matched mixtures. D1--D5 denote the matched Gaussian--Poisson conditions ordered from the weakest to the strongest noise.\par

\begin{figure}[!htbp]
\centering
\includegraphics[width=0.98\columnwidth,keepaspectratio]{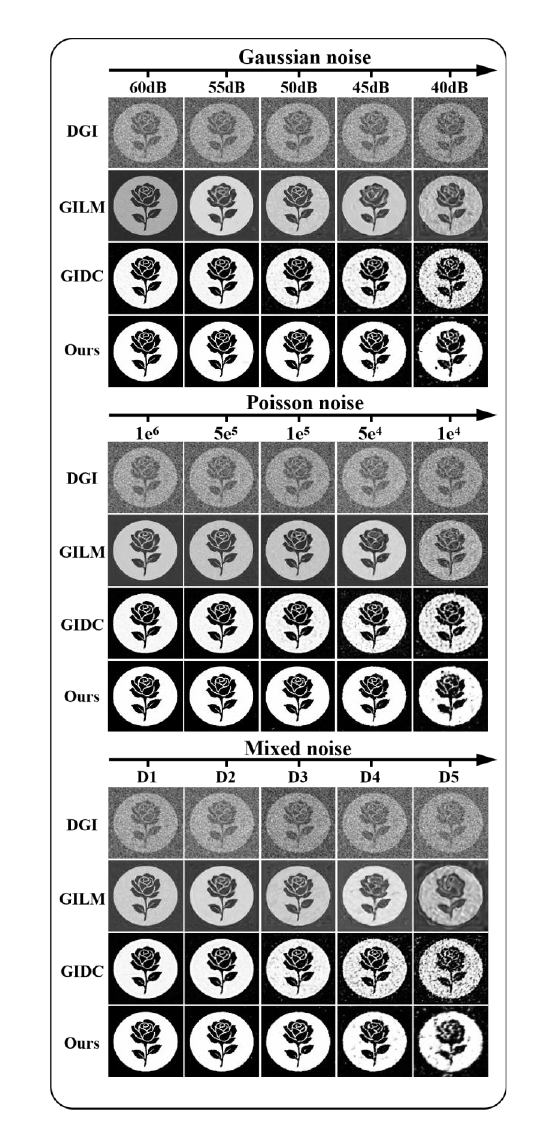}
\caption{Reconstruction results under Gaussian, Poisson, and mixed noise. Noise strength increases from left to right.}
\label{fig:noise-robustness}
\end{figure}

\begin{table}[!htbp]
\centering
\scriptsize
\caption{Learnable thresholds under different noise levels.}
\label{tab:learnable-threshold}
\begin{tabular*}{\columnwidth}{@{\extracolsep{\fill}}lccccc@{}}
\toprule
\multicolumn{6}{c}{\textbf{Gaussian noise}} \\
\midrule
Level & 60 dB & 55 dB & 50 dB & 45 dB & 40 dB \\
Threshold & 0.0345 & 0.0353 & 0.0391 & 0.0412 & 0.0472 \\
\midrule
\multicolumn{6}{c}{\textbf{Poisson noise}} \\
\midrule
Level & $1\times10^6$ & $5\times10^5$ & $1\times10^5$ & $5\times10^4$ & $1\times10^4$ \\
Threshold & 0.0368 & 0.0389 & 0.0385 & 0.0403 & 0.0465 \\
\midrule
\multicolumn{6}{c}{\textbf{Mixed noise}} \\
\midrule
Level & \shortstack{D1} & \shortstack{D2} & \shortstack{D3} & \shortstack{D4} & \shortstack{D5} \\
Threshold & 0.0391 & 0.0412 & 0.0421 & 0.0452 & 0.0492 \\
\bottomrule
\end{tabular*}
\end{table}

The visual comparisons in Fig.~\ref{fig:noise-robustness} show that SISTA-Net preserves clearer target structures and suppresses background noise more effectively as the noise strength increases, particularly under strong mixed noise. Table~\ref{tab:learnable-threshold} reports the learned threshold corresponding to the final reconstruction selected by the early-stopping strategy at each noise level. The thresholds exhibit an overall upward trend as the noise strength increases, indicating that the adaptive shrinkage becomes stronger for noisier measurements. This response suppresses noise-induced weak coefficients while retaining dominant structural features, thereby contributing to robust reconstruction.\par
\vspace*{2mm} \noindent \textbf{4\quad Underwater Environment Experiments}\vspace*{2mm}

\vspace*{2mm}\noindent\emph{4.1\quad Experimental setup}\vspace*{2mm}

\begin{figure}[!h]
\centering
\includegraphics[width=1\columnwidth,keepaspectratio]{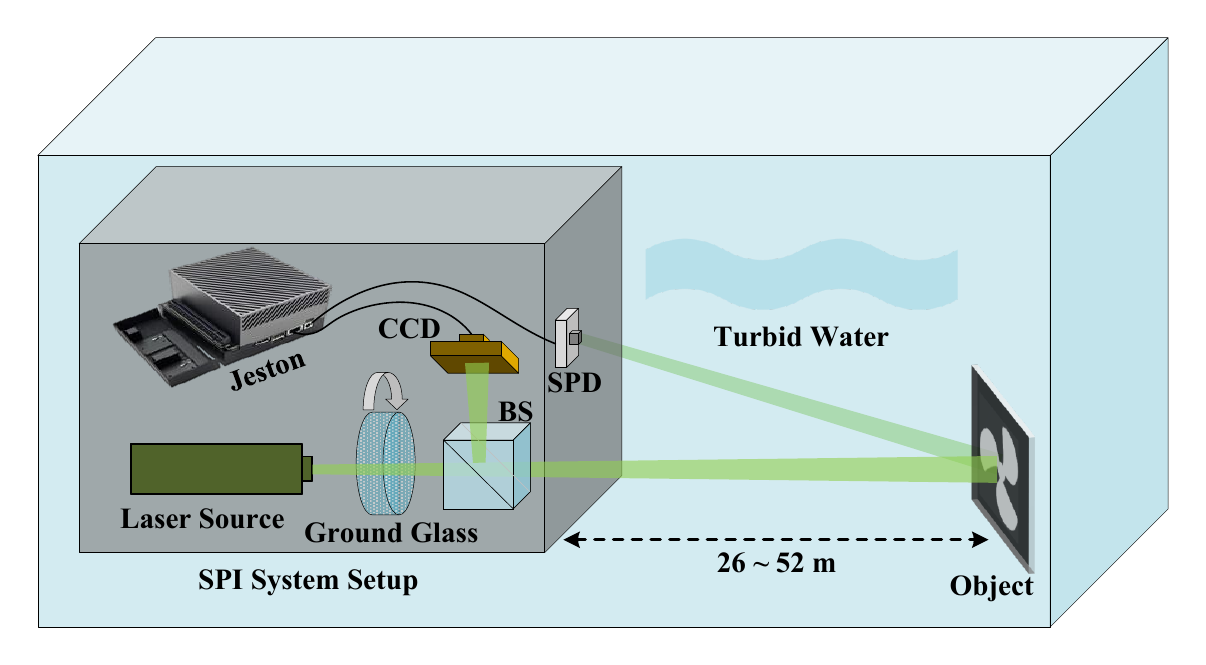}
\caption{ Experimental setup for underwater SPI.}
\label{fig:setup}
\end{figure}

\begin{figure}[!tb]
\centering
\includegraphics[width=0.98\columnwidth,height=0.735\textheight,keepaspectratio]{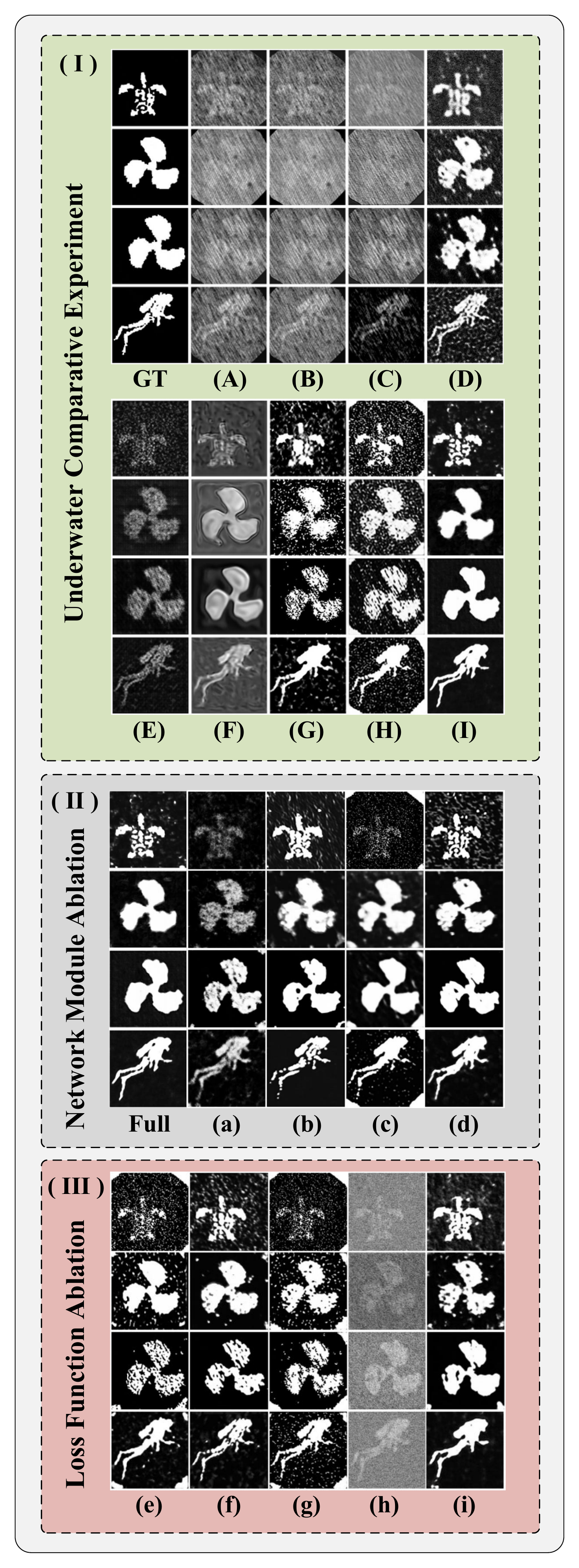}
\caption{Reconstruction and ablation results.(I) Reconstruction results. (II) Network module ablation. (III) Loss function ablation.}
\label{fig:real-ab}
\end{figure}

\begin{table*}[!tbp]
\centering
\scriptsize
\caption{Quantitative underwater reconstruction results.}
\label{tab:real-data}
\resizebox{\textwidth}{!}{%
\begin{tabular}{lccccccccccc}
\toprule
\multirow{2}{*}{Method} & \multirow{2}{*}{Params(M)} & \multirow{2}{*}{GFLOPs} & \multirow{2}{*}{Time(s)} & \multicolumn{2}{c}{frogman(f-52)} & \multicolumn{2}{c}{propeller50(p-50)} & \multicolumn{2}{c}{propeller52(p-52)} & \multicolumn{2}{c}{tortoise(t-26)} \\
\cmidrule(lr){5-6}\cmidrule(lr){7-8}\cmidrule(lr){9-10}\cmidrule(lr){11-12}
    &  &  &  & PSNR & CNR & PSNR & CNR & PSNR & CNR & PSNR & CNR \\
\midrule
\textbf{DGI} & -- & -- & -- & 7.95 & 1.7421 & 7.15 & 1.2625 & 6.24 & 0.8519 & 7.80 & 1.7309 \\
\textbf{ISTA-Net++} & 0.34 & 13.32 & 8.46 & 7.29 & 1.4764 & 6.92 & 1.1044 & 5.80 & 0.7432 & 7.05 & 1.4999 \\
\textbf{TransCS} & 1.76 & 65.46 & 9.04 & 10.92 & 2.7050 & 7.34 & 1.2229 & 6.22 & 0.6739 & 5.75 & 1.1028 \\
\textbf{Restormer} & 26.12 & 310.24 & 13.76 & 11.31 & 3.1662 & 11.90 & 3.4370 & 10.80 & 3.0079 & 9.43 & 2.6061 \\
\textbf{URNet} & 61.22 & 14.57 & 2.64 & 12.23 & 6.7178 & 10.09 & 5.4472 & 10.18 & 5.5894 & 12.35 & 3.2188 \\
\textbf{GILM} & 109.50 & 10646.54 & 16.19 & 10.17 & 6.4421 & 12.75 & 5.4308 & 9.53 & 4.7341 & 10.83 & 4.6355 \\
\textbf{UNet} & 26.18 & 67.21 & 2.19 & 14.80 & 5.2798 & 9.37 & 4.0466 & 9.37 & 2.3993 & 11.96 & 3.5760 \\
\textbf{GIDC} & 6.55 & 16.83 & 1.84 & 10.75 & 2.9317 & 9.44 & 2.2283 & 9.59 & 2.4095 & 9.50 & 2.4464 \\
\textbf{Ours} & 3.18 & 264.10 & 10.28 & \textbf{15.48} & \textbf{6.9405} & \textbf{14.44} & \textbf{5.6981} & \textbf{15.57} & \textbf{5.9008} & \textbf{13.40} & \textbf{4.7068} \\
\bottomrule
\end{tabular}%
}
\end{table*}

\noindent To validate the proposed method beyond simulations and under real-world perturbations, we deployed an underwater SPI platform, as shown in Fig.~\ref{fig:setup}. A laser provided active illumination, which was modulated by a rotating ground glass to generate random speckle patterns. A beam splitter (BS) divided the modulated light: one path illuminated the underwater target, with the reflected 1D measurement signals collected by a single-photon detector (SPD) serving as the self-supervisory signals; the other path was recorded by a charge-coupled device (CCD) to capture the 2D pattern distributions.The real-world tests included four far-field targets: a sea turtle model at 26 m, a propeller board at 50 m, and separate propeller and frogman boards at 52 m. For underwater reconstruction without ground truth, we use a measurement-consistency plateau criterion, stopping optimization when the residual between the measured and forward-projected signals remains stable.\par

\vspace*{2mm}\noindent\emph{4.2\quad Results}\vspace*{2mm}

\noindent The reconstruction results in Fig.~\ref{fig:real-ab}(I) demonstrate the robustness of SISTA-Net in practical underwater SPI scenarios. Under scattering and detector noise, the real measurements are strongly degraded and corrupted by interference\cite{degradations-for-GI}. The correlation-based baseline produces noisy and low-contrast reconstructions, while the learning-based baselines still suffer from residual artifacts or blurred boundaries. In contrast, SISTA-Net reconstructs clearer target structures with sharper boundaries and cleaner background regions.\par

The quantitative results in Table~\ref{tab:real-data} further confirm this visual observation. Since ground-truth images are unavailable in underwater experiments, a high-fidelity pseudo ground truth was constructed through background subtraction, Total Variation regularization, BM3D denoising, and morphological operations. SISTA-Net achieves the best PSNR and Contrast-to-Noise Ratio (CNR) on all real underwater scenes, indicating superior reconstruction quality and anti-interference capability. Here, CNR measures the target contrast relative to the background noise. For practical efficiency, we also report Params, GFLOPs, and average reconstruction time. Although SISTA-Net has a moderate parameter size, its GFLOPs and reconstruction time are relatively high, mainly because the VSSM module unfolds 2D features into directional sequences, introducing memory rearrangement overhead, and performs linear projections and selective scans along multiple directions. Therefore, the current design favors reconstruction robustness and quality over reconstruction speed, and further acceleration of the state-space module will be considered in future work.\par

\begin{table*}[!htbp]
\centering
\caption{PSNR results for ablation study.}
\label{tab:psnr-ablation}
\begingroup
\renewcommand{\arraystretch}{1.12}
\scriptsize
\newcommand{\graycell}{\cellcolor{gray!20}}
\resizebox{\textwidth}{!}{%
\begin{tabular}{lcccccccccccc}
\toprule
\multirow{2}{*}{Method} & \multicolumn{4}{c}{Network Module Ablation} & \multicolumn{4}{c}{Loss Function Ablation} & \multicolumn{4}{c}{PSNR (dB)} \\
\cmidrule(lr){2-5}\cmidrule(lr){6-9}\cmidrule(lr){10-13}
    & Proximal & Res2MM & Win-F & Win-P & $L_{F}$ & $L_{S}$ & $L_{P}^{KL}$ & $L_{P}^{MSE}$ & f-52 & p-52 & p-50 & t-26 \\
\midrule
(a) &  & $\surd$ & $\surd$ &  & \graycell & \graycell & \graycell & \graycell & 14.87 & 12.83 & 12.12 & 13.03 \\
(b) & $\surd$ & $\surd$ & $\surd$ &  & \graycell & \graycell & \graycell & \graycell & 14.86 & 13.01 & 14.68 & 12.12 \\
(c) & $\surd$ &  &  & $\surd$ & \graycell & \graycell & \graycell & \graycell & 11.63 & 12.06 & 13.27 & 9.88 \\
(d) & $\surd$ & $\surd$ &  &  & \graycell & \graycell & \graycell & \graycell & 15.46 & 13.65 & 14.76 & 12.12 \\
(e) & \graycell & \graycell & \graycell & \graycell &  & $\surd$ & $\surd$ &  & 12.38 & 9.46 & 12.02 & 8.76 \\
(f) & \graycell & \graycell & \graycell & \graycell & $\surd$ &  & $\surd$ &  & 15.25 & 12.03 & 14.03 & 11.82 \\
(g) & \graycell & \graycell & \graycell & \graycell &  &  & $\surd$ &  & 11.00 & 8.98 & 9.89 & 8.91 \\
(h) & \graycell & \graycell & \graycell & \graycell & $\surd$ & $\surd$ &  &  & 5.46 & 5.76 & 6.29 & 5.18 \\
(i) & \graycell & \graycell & \graycell & \graycell & $\surd$ & $\surd$ &  & $\surd$ & 14.69 & 12.05 & 13.26 & 11.32 \\
\midrule
{\bfseries\boldmath Full} & {\bfseries\boldmath $\surd$} & {\bfseries\boldmath $\surd$} & {\bfseries\boldmath $\surd$} & {\bfseries\boldmath $\surd$} & {\bfseries\boldmath $\surd$} & {\bfseries\boldmath $\surd$} & {\bfseries\boldmath $\surd$} &  & {\bfseries\boldmath 15.48} & {\bfseries\boldmath 14.44} & {\bfseries\boldmath 15.57} & {\bfseries\boldmath 13.4} \\
\bottomrule
\end{tabular}%
}
\endgroup
\end{table*}

\vspace*{2mm}\noindent\emph{4.3\quad Ablation Study}\vspace*{2mm}

\noindent To rigorously validate the necessity of SISTA-Net's core architecture and constraint mechanisms, we conducted systematic ablation studies under identical experimental settings. By designing model variants across two main dimensions, namely network architecture and loss functions, we individually evaluated the specific contributions of each mechanism.\par

Regarding the network architecture ablation, we designed four variants: (a) removing the proximal mapping module and retain only the fidelity module to confirm the core role of the sparsity constraint; (b) replacing the WMB in the proximal mapping module with pure convolutions to verify the advantage of state space models in extracting deep sparse features; (c) substituting the hybrid fidelity architecture with a standard UNet to demonstrate the high adaptability of our customized Res2MM-Net to SPI reconstruction and the plug-and-play versatility of the proximal mapping network; (d) discarding local window partitioning and adopt global scanning to reveal the critical role of local spatial modeling.\par

Regarding the loss function ablation, we designed five variants: (e) removing the fidelity loss to validate its role in aligning reconstruction results with physical measurements; (f) removing the $L_1$ sparsity loss to verify its function in enforcing feature sparsity and suppressing redundant information in the latent domain; (g) removing both the fidelity loss and sparsity loss simultaneously and retaining only the proximal response constraint; (h) removing the KL-based proximal loss to directly evaluate its contribution; (i) replacing the KL-based proximal response loss with an additional proximal measurement-domain MSE loss to examine whether the KL term is redundant with the fidelity MSE.\par

As illustrated in Fig.~\ref{fig:real-ab}(II) and Fig.~\ref{fig:real-ab}(III), all variant models exhibited varying degrees of reconstruction performance degradation, validating the indispensable and tightly-coupled synergistic optimization of core framework components. The quantitative PSNR results are presented in Table~\ref{tab:psnr-ablation}. 

\noindent For network architecture, removing the proximal mapping module confirmed its core role in sparse constraint enforcement, replacing WMB with pure convolutions verified the superiority of state space models in extracting deep sparse features, substituting the hybrid fidelity architecture with standard UNet demonstrated Res2MM-Net's high adaptability to SPI reconstruction and the proximal mapping network's plug-and-play versatility, and discarding local window partitioning highlighted the critical value of local spatial modeling. 
1

 For loss functions, removing the fidelity loss caused misalignment between reconstruction results and physical measurements, removing the $L_1$ sparsity loss weakened feature sparsity and failed to suppress latent-domain redundant information, and retaining only the proximal response constraint exacerbated underfitting and structural degradation. Removing the KL-based proximal loss led to severe performance degradation, confirming its stabilizing role for the final proximal output. Replacing it with a proximal MSE loss improved over removing KL but remained inferior to the full model, indicating that the KL term is not a simple duplicate of the measurement MSE but provides complementary relative-response consistency. These results confirm that the synergy between architectural modules and complementary loss constraints is essential for filtering redundant noise and ensuring stable and efficient reconstruction in self-supervised paradigm.

\vspace*{2mm} \noindent \textbf{5\quad Conclusion}\vspace*{2mm}

\noindent We propose SISTA-Net, which effectively resolves the severe reliance on paired datasets and the lack of physical interpretability, while achieving excellent noise suppression. Motivated by ISTA, we construct a physically interpretable architecture consisting of a data fidelity module and a proximal mapping module. By integrating VSSM-driven global dependencies with convolutional local details, Res2MM-Net ensures robust feature fidelity, while the proximal mapping module enforces $L_1$ sparsity and physical consistency to achieve precise noise filtering in the latent domain. Extensive experiments demonstrate that the proposed method achieves high-quality reconstruction relying solely on 1D physical measurements. Compared to mainstream baselines, our method not only exhibits significant advantages in detail preservation under extremely low sampling ratios but also demonstrates robust anti-interference and generalization capabilities in complex real-world underwater environments. In summary, this study successfully bridges the gap between traditional physics-based optimization and self-supervised learning, offering a highly promising solution for low-cost, high-fidelity SPI in real-world applications.\par

\vspace*{2mm} \noindent \emph{Acknowledgments} \vspace*{2mm}

\noindent This work was supported by Fujian Ocean Innovation Center (Grant No.25FV0CZJ09).\par

{\small
\bibliographystyle{unsrt}
\bibliography{reference/reference}
}

\end{document}